\documentclass{article}

\usepackage[final]{nips_2017}

\usepackage[utf8]{inputenc} % allow utf-8 input
\usepackage[T1]{fontenc}    % use 8-bit T1 fonts
\usepackage{hyperref}       % hyperlinks
\usepackage{url}            % simple URL typesetting
\usepackage{booktabs}       % professional-quality tables
\usepackage{amsfonts}       % blackboard math symbols
\usepackage{nicefrac}       % compact symbols for 1/2, etc.
\usepackage{microtype}      % microtypography
\usepackage{comment}
\usepackage{graphicx}
\usepackage{chngpage}

\title{CliNER 2.0: Accessible and Accurate\\Clinical Concept Extraction}

\author{
  Willie Boag\\
  MIT CSAIL\\
  Cambridge, MA\\
  \texttt{wboag@mit.edu}\\
  \And
  Elena Sergeeva\\
  MIT CSAIL\\
  Cambridge, MA\\
  \texttt{elenaser@mit.edu}\\
  \And
  Saurabh Kulshreshtha\\
  UMass Lowell\\
  Lowell, MA\\
  \texttt{skul@cs.uml.edu}\\
  \AND
  Peter Szolovits\\
  MIT CSAIL\\
  Cambridge, MA\\
  \texttt{psz@mit.edu}\\
  \And
  Anna Rumshisky\\
  UMass Lowell\\
  Lowell, MA\\
  \texttt{arum@cs.uml.edu}\\
  \And
  Tristan Naumann\\
  MIT CSAIL\\
  Cambridge, MA\\
  \texttt{tjn@mit.edu}\\
}

\begin{document}

\maketitle

%%%%%%%%%%%%%%%%%%%%%%%%%%%%%%%%%%%%%%%%%%%%%%%%%%%%%%%%%%%%%%%%
% Abstract
%%%%%%%%%%%%%%%%%%%%%%%%%%%%%%%%%%%%%%%%%%%%%%%%%%%%%%%%%%%%%%%%
\begin{abstract}
% Clinical concept extraction (CCE) is important for understanding clinical notes and provides a foundation for many downstream clinical decision-making tasks. Notes often describe the most important aspects of a patient's stay and are therefore critical to medical research; one of the most important first steps in reasoning involves identifying the named entities in the notes: problems, tests, and treatments. 
% %
% In the past, this task has been posed as a standard NER sequence tagging problem, solved with feature-based methods with hand-engineered domain knowledge. Recent advances, however, have shown the effectiveness of LSTM-based models for NER tasks, including CCE.
% %
% This work presents \textit{CliNER 2.0}, a simple-to-install, open-source tool for extracting concepts from clinical text. \textit{CliNER 2.0} uses a word- and character- level LSTM model, and achieves state-of-the-art performance. For ease of use, the tool also includes pre-trained models available for public use.

% Tasks such as the 2010 i2b2/va shared workshop have made important steps forward in releasing data to build models that can identify important clinical concepts, however the competition format has not incentivized open and easy-to-use tools for researchers that are looking for simple concept extraction solutions. 

Clinical notes often describe important aspects of a patient's stay and are therefore critical to medical research. Clinical concept extraction (CCE) of named entities --- such as problems, tests, and treatments --- aids in forming an understanding of notes and provides a foundation for many downstream clinical decision-making tasks. Historically, this task has been posed as a standard named entity recognition (NER) sequence tagging problem, and solved with feature-based methods using hand-engineered domain knowledge. Recent advances, however, have demonstrated the efficacy of LSTM-based models for NER tasks, including CCE. This work presents \textit{CliNER 2.0}, a simple-to-install, open-source tool for extracting concepts from clinical text. CliNER 2.0 uses a word- and character- level LSTM model, and achieves state-of-the-art performance. For ease of use, the tool also includes pre-trained models available for public use.

\end{abstract}

%%%%%%%%%%%%%%%%%%%%%%%%%%%%%%%%%%%%%%%%%%%%%%%%%%%%%%%%%%%%%%%%
% Intro
%%%%%%%%%%%%%%%%%%%%%%%%%%%%%%%%%%%%%%%%%%%%%%%%%%%%%%%%%%%%%%%%
\section{Introduction}

% SECONDARY USE OF EHR
Although there is a trend toward digitizing patient records in an increasingly structured manner, much information is still hidden in unstructured narrative text. In their primary role, electronic health record (EHR) notes facilitate patient care by recording communication among care staff. These clinical notes capture patient data that provide insight into a patient's status and courses of care, such as patient history, recommended treatments, records of meetings, and more. Often, this granularity of data does not appear, or does not appear in equivalent detail, in a structured form elsewhere in the EHR. It is no surprise then, that leveraging clinical notes is critical to the successful analysis of EHR data --- an important secondary use.

\begin{figure}[t]
 \centering
 \includegraphics[width=90mm]{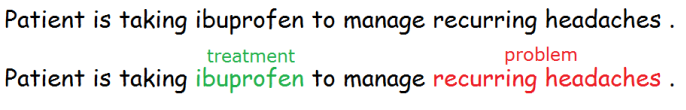}
%  \vspace{-1em}
 \caption{Identifying concept spans in clinical text. Concept spans can be any number of contiguous tokens; in this case there are spans of length one and two, respectively.}
 \label{iob-tagging}
\end{figure}

% DESCRIPTION OF TASK
% In order to perform reasoning and analysis of the notes, we must extract the important concepts: 
Clinical concept extraction (CCE) improves our understanding of notes and our ability to analyze them; identifying for example,
\textit{problems} and symptoms a patient has exhibited, \textit{tests} that have been run, and \textit{treatments} that have been administered. 
CCE, much like standard named entity recognition (NER), has traditionally been posed as a sequence tagging task, handled by feature-based methods using hand-engineered domain knowledge. In this formulation, tokens (i.e., pre-processed words) are predicted to be inside, outside, or beginning (IOB) a concept span; thus, allowing the identification concepts that span multiple, contiguous tokens. Subsequently, each identified span is predicted to be one of the specified concepts, as shown in Figure~\ref{iob-tagging}.

% COMMUNITY EFFORTS
Community efforts to advance concept extraction have led to numerous shared task competitions. Notably, in 2010 Informatics for Integrating Biology \& the Bedside (i2b2) and the U.S. Department of Veterans Affairs (VA) partnered to hold the Workshop on Natural Language Processing Challenges for Clinical Records~\citep{uzuner:i2b2}. This workshop included a task for concept extraction from clinical discharge summaries, the objective of which was to identify contiguous spans of tokens as in narrative text and classify their concept type.
% Moved to caption
%Concept spans could be any number of contiguous tokens, in the case of Figure \ref{iob-tagging}, there are spans of length one and two, respectively.

The 2010 i2b2/VA workshop made an important step forward in releasing the data required for building tools to identify clinically important concepts.
% NOTE: This link is now provided in the Data section below.
%~\footnote{This data can be found at \url{https://www.i2b2.org/NLP/DataSets/Main.php}.} 
However, despite the existence of tasks like this, the competition format has not incentivized open and easy-to-use tools for researchers who require simple solutions. Most submissions are not released as open systems, thus limiting the availability of concept extraction tools. To address these concerns, the CliNER project was developed as an open-source clone of a top-performing i2b2/VA 2010 challenge system \cite{boag:cliner}. 
% TODO: This model was meh. We have improved it.

% CLINER TO THE RESCUE!
This work presents \textit{CliNER 2.0}, a simple-to-install, open-source tool for extracting concepts from clinical text, which includes pre-trained models for additional ease of use.~\footnote{CliNER 2.0 can be downloaded from \url{https://github.com/text-machine-lab/CliNER}.} Much like general domain natural language processing (NLP), clinical NLP has also been shown to benefit from deep learning models that can better learn complex patterns from data. 
Recently, \citet{dernoncourt2016identification} proposed a word- and character-level LSTM model for the de-identification task that outperformed all existing baselines. 
In \textit{CliNER 2.0}, we adopt this approach for concept extraction, integrating the same state-of-the-art deep learning NER architecture described above into the tool.

\section{Related Work}

The 2010 i2b2/VA Workshop on NLP Challenges for Clinical Records~\citep{uzuner:i2b2} promoted the development of 22 systems towards the task of concept extraction from discharge summaries. 
%The first of this workshop's three tasks was concept extraction from discharge summaries, for which 22 systems were developed. 
The winning system achieved an exact F measure of 0.852 by using a discriminative semi-Markov HMM, trained using passive-aggressive online updates~\citep{deBruijn}. Many other top performing methods used a Conditional Random Field (CRF) to model the sequence learning problem~\citep{Harabagiu}.

In the years following the shared task workshop, the dataset proved very useful as a research benchmark. Numerous systems and methods that have been developed can be compared against one another using this dataset.
Early successful attempts utilized the strengths of workshop participants (sequential models, such as a CRF) and added generalized word representations using distributional semantics~\citep{fu2014improving,jonnalagadda2012enhancing,wu2015:neural}. Since then, deep learning and recurrent neural networks have increased in popularity and easiness-to-implement, leading to a many LSTM-based approaches to clinical concept extraction~\citep{chalapathy2016bidirectional,cce-wc-lstm2017}.

The most widely-used clinical NLP tool, cTAKES~\citep{savova2010ctakes}, relies nearly exclusively on UMLS-based dictionary lookups~\citep{bodenreider2004umls}.
In doing so, cTAKES achieves high recall (at the cost of low precision) by identifying all phrases that have any potential to be a relevant concept.
While this property may be desirable for search-related tasks, it's lack of relevance to many downstream clinical decision-making tasks has been noted as the reason for the development of additional tools~\citep{divita2014sophia,kang2017eliie,soysal2017clamp}. This limitation, in addition to a desire for out-of-the-box usability motivated the creation of the original CliNER~\citep{boag:cliner}
\section{Data}

We use data from the i2b2/VA 2010 challenge.~\footnote{i2b2/VA 2010 challenge data are available at \url{https://www.i2b2.org/NLP/DataSets/Main.php}.} This dataset contains 16,107 sentences, which are 6--9 words long on average, from patient discharge summaries. There are 169 summaries made available for training and 255 summaries available for testing. Note the training data are smaller than the testing data, a result of nearly a third of the training data being revoked following the challenge.

%\textbf{TODO: DESCRIBE HOW THE THIRD HOSPITAL DATA WAS REDACTED AND WHY THERE IS LESS TRAINING DATA.}

%%%%%%%%%%%%%%%%%%%%%%%%%%%%%%%%%%%%%%%%%%%%%%%%%%%%%%%%%%%%%%%%
% Methods
%%%%%%%%%%%%%%%%%%%%%%%%%%%%%%%%%%%%%%%%%%%%%%%%%%%%%%%%%%%%%%%%
\section{Methods}
CliNER 2.0 has two options for building machine learning models: 
1) traditional CRF-based learning with domain expert features, or 
2) deep learning with a state-of-the-art neural architecture. These options afford flexibility with respect to desired time and space constraints; notably, the CRF model is smaller than a large hierarchical LSTM. Both models employ a word-level prediction using the IOB format. For three concept types --- problem, test, and treatment --- this results in a 7-way tag prediction for each token.

\subsection{CRF with UMLS Features}
The CRF-based option heavily employs feature extraction using both linguistic features (e.g., ngrams and wordshapes) and domain knowledge (e.g., UMLS Metathesaurus). POS tagging was performed with the general-domain nltk pos\_tagger \citep{nltk}. Table~\ref{tab:features} shows a full list of features that are extracted for each token. These features are extracted for each individual token, except for \textit{prev1-all-feats} and \textit{next1-all-feats}, which include all word-level features of the previous and next tokens, respectively. These features are then fed into a wrapper for CRFsuite, a fast CRF implementation~\citep{CRFsuite}.

\begin{table*}[h]
\begin{center}
\caption{Features for the CRF.}
\label{tab:features}
 
	\begin{tabular}{|l|l|l|l|}
      \hline
word unigram     & last-2 characters &  word shape      & part-of-speech \\ \hline
regexes of units &  length           &  umls-cui        & umls-lui \\ \hline
umls-rel         &   umls-sty        &   umls-tty       & umls-abr  \\ \hline
prev3-unigrams   &  next3-unigrams   &  \textit{prev1-all-feats} & \textit{next1-all-feats} \\ %PREV1-ALL-FEATS & NEXT1-ALL-FEATS \\
\hline 
   \end{tabular}
\end{center}
\end{table*}

\subsection{Hierarchical LSTM}

\begin{figure}[h]
\centering 
 \includegraphics[scale=0.5]{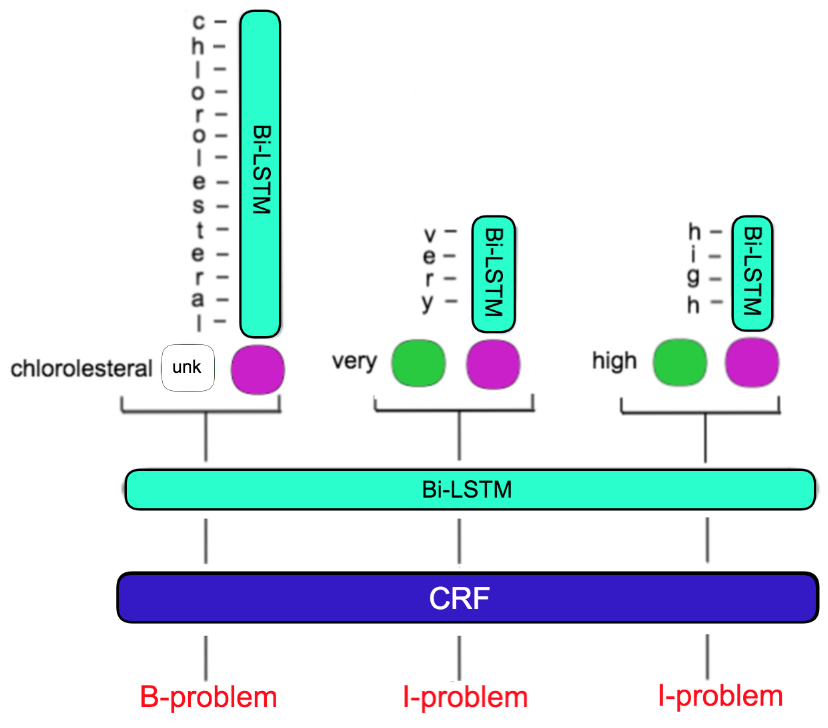}
 \caption{Character-level LSTMs that feed into word-level LSTMs. Unlike purely word-level models, this approach is sensitive to misspellings such as ``cholorolesteral.''}
 \label{fig:wc-lstm}
\end{figure}

The hierarchical LSTM option employs both word- and character- level bidirectional LSTMs (w+c Bi-LSTM). For each word $w_t$ in a sentence, we consider the sequence of characters that comprise that word: $w_t^1$, $w_t^2$, $w_t^3, ...$. The embeddings for this sequence of characters $w_t^i$ are fed into the Bi-LSTM$_t$ corresponding to the $t^{th}$ word, and concatenated to the final forward and backward hidden states to create a character-level representation of the word. All character-level Bi-LSTMs share the same weights. Finally, the character-level representation is concatenated with its standard word embedding (pre-trained GloVe vectors available at \url{http://neuroner.com/data/word_vectors/glove.6B.100d.zip}) to form a rich word- and character- level representation of token $t$. From there, this word representation is fed in to a standard (Bi-)LSTM-CRF framework. Figure~\ref{fig:wc-lstm} depicts the w+c LSTM-CRF model.

%%%%%%%%%%%%%%%%%%%%%%%%%%%%%%%%%%%%%%%%%%%%%%%%%%%%%%%%%%%%%%%%
% Results
%%%%%%%%%%%%%%%%%%%%%%%%%%%%%%%%%%%%%%%%%%%%%%%%%%%%%%%%%%%%%%%%
\section{Results}

Table~\ref{tab:concepts} displays the results from recent work on the 2010 i2b2/VA concept extraction task. Notably excluded are the results of the top-performing systems from the original task. These systems are removed because their reported performance was obtained using now-revoked training data and the systems are not available to train again using the limited subset. Consequently, their reported performances from the 2010 competition are not comparable to recent work.

The remaining top-performing systems are are all deep neural models using LSTMs and neural embeddings. Their relative performances vary and no single, clear winner stands out. While the Binarized Neural Embedding CRF achieves the best precision, the other LSTM-CRF models (some of which also use character-based LSTMs) all independently achieve the best recall and F1 scores. 

The nearly identical performance of three systems --- and the limited gains of additional character LSTMs, pre-trained word embeddings, and other enhancements --- suggest that the community might be reaching the limits of this task. Performance on the (rather small) i2b2 dataset has effectively plateaued, with little-to-no recent improvements in performance.

Notably, both the shallow CRF and deep w+c LSTM-CRF models available in CliNER 2.0 match, or exceed, the top performing systems in their respective machine learning paradigm. This puts CliNER 2.0 performance among the highest possible using either of the methods.

\begin{table*}[t]
\begin{center}
\caption{Precision, recall, and F1 of selected concept extraction models.}
\label{tab:concepts}
	\begin{tabular}{|l|l|l|l|}
      \hline
      & \multicolumn{3}{|c|}{Exact Class Match}\\ \hline
                     &  Precision     &  Recall         &  F1     \\ \hline
% Truecasing CRFSuite (Fu and Ananiadou, 2014)            &         0.808  & 0.715 & 0.759   \\ \hline
% Binarized Neural Embedding CRF (Wu et al., 2015)        &         \textbf{0.851} & 0.806 & 0.828 \\ \hline
% LSTM-CRF: GloVe     (Chalapathy et al. 2016)            &         0.844 & 0.834 & \textbf{0.839}   \\ \hline
% w+c LSTM-CRF: CommonCrawl (Unanue et al. 2017)          & ---- & ---- & 0.834 \\
Truecasing CRFSuite \citep{fu2014improving}             &         0.808  & 0.715 & 0.759   \\ \hline
Binarized Neural Embedding CRF \citep{wu2015:neural}    &         \textbf{0.851} & 0.806 & 0.828 \\ \hline
LSTM-CRF: GloVe     \cite{chalapathy2016bidirectional}  &         0.844 & 0.834 & \textbf{0.839}   \\ \hline
%w+c LSTM-CRF: GloVe    \cite{lample2016neural}  &         0.799 & 0.836 & 0.818   \\ \hline
%w+c LSTM-CRF: CommonCrawl \citep{cce-wc-lstm2017}       & ---- & ---- & 0.834 \\ \hline
\hline
   CliNER 2.0: feats+CRF                                &         0.835  &         0.758  & 0.795   \\ \hline
   CliNER 2.0: w+c LSTM-CRF: GloVe                      &         0.840  &         \textbf{0.836}  & 0.838   \\ \hline	
   \end{tabular}
\end{center}
\end{table*}

%%%%%%%%%%%%%%%%%%%%%%%%%%%%%%%%%%%%%%%%%%%%%%%%%%%%%%%%%%%%%%%%
% Conclusion
%%%%%%%%%%%%%%%%%%%%%%%%%%%%%%%%%%%%%%%%%%%%%%%%%%%%%%%%%%%%%%%%
\section{Conclusion}
We present \textit{CliNER 2.0}, an updated, open-source clinical named entity recognition tool that matches state-of-the-art performance, and achieves the highest reported recall among systems trained on the 2010 i2b2 data. 
%While other tools like cTakes have very high recall, their false positive rates result in an overwhelming number of concept spans being identified, most of which are not relevant. 
While other tools like cTakes often identify a large number possible concepts for a given span, it can be overwhelming when many are not relevant.
\textit{CliNER 2.0}, on the other hand, has a much less intrusive number of false positives, and focuses specifically on the identification of relevant concepts: problems, tests, and treatments.

Further, \textit{CliNER 2.0} is easy to install. Pre-trained models are available for public use, which allow the tool to run out-of-the-box. To our knowledge, this represents the first open-source and pre-trained deep learning model available for state-of-the-art concept extraction. In addition, the tool has an optional flag that can disable the deep network; instead, backing off to a simple CRF model with UMLS-derived features. This option could be useful in resource constrained settings.
%This option could be useful on machines with performance constraints.

%%%%%%%%%%%%%%%%%%%%%%%%%%%%%%%%%%%%%%%%%%%%%%%%%%%%%%%%%%%%%%%%
% Acknowledgements
%%%%%%%%%%%%%%%%%%%%%%%%%%%%%%%%%%%%%%%%%%%%%%%%%%%%%%%%%%%%%%%%
\section*{Acknowledgments}
% REDACTED FOR REVIEW
% TODO: for each author, make sure funding is acknowledged.
% [ ] TODO(arum)
% [ ] TODO(elenaser)
% [ ] TODO(skul)
% [x] TODO(tjn)
% [x] TODO(wboag)

This research was funded in part by
the Intel Science and Technology Center for Big Data,
a Philips-MIT research agreement,
the National Science Foundation Graduate Research Fellowship Program grant No. 1122374,
and grants from the National Institutes of Health (NIH):
National Library of Medicine (NLM) Biomedical Informatics Research Training grant 2T15 LM007092-22.

% This research was funded in part by the Intel Science and Technology Center for Big Data, the National Library of Medicine Biomedical Informatics Research Training grant (NIH/NLM 2T15 LM007092-22).

% This material is based upon work supported by the National Science Foundation Graduate Research Fellowship Program under Grant No. 1122374. Any opinions, findings, and conclusions or recommendations expressed in this material are those of the author(s) and do not necessarily reflect the views of the National Science Foundation.

% \newpage
\small
\bibliographystyle{abbrvnat}
\bibliography{nips}

\end{document}